\documentclass[final,1p,times]{elsarticle}

\usepackage{amsmath}
\usepackage{subfigure}
\usepackage{multirow}
\usepackage{url}
\usepackage{booktabs}

\journal{Data Intelligence}

\begin{document}

\begin{frontmatter}

\title{Deep Learning with Heterogeneous Graph Embeddings for Mortality Prediction from Electronic Health Records}

\author[Sinai HPIMS,Indiana]{Tingyi Wanyan} %\ead{tingyi.wanyan@mssm.edu}
   \author[Sinai HPIMS]{Hossein Honarvar} %\ead{hossein.honarvargheitanbaf@mssm.edu}
   \author[Indiana]{Ariful Azad} %\ead{azad@iu.edu}
   \author[UT Austin Dell,UT Austin In]{Ying Ding} %\ead{ying.ding@austin.utexas.edu}
   \author[Sinai HPIMS,Sinai GGS]{Benjamin S. Glicksberg*}
\cortext[Ben details]{Corresponding author}

\ead{benjamin.glicksberg@mssm.edu}
\address{770 Lexington Ave., 14th Floor,
New York, NY 10029
(212) 731-7078}

%Affiliations
\address[Sinai HPIMS]{Hasso Plattner Institute for Digital Health at Mount Sinai, Icahn School of Medicine at Mount Sinai, New York, NY, USA.}
\address[Indiana]{School of Informatics, Computing, and Engineering, Indiana University, Bloomington, IN, USA.}
\address[UT Austin Dell]{Dell Medical School, University of Texas at Austin, Austin, TX, USA.}
\address[UT Austin In]{School of Informatics, University of Texas at Austin, Austin, TX, USA.}
\address[Sinai GGS]{Department of Genetics and Genomic Sciences, Icahn School of Medicine at Mount Sinai, New York, NY, USA.}

\begin{abstract}
Computational prediction of in-hospital mortality in the setting of an intensive care unit can help clinical practitioners to guide care and make early decisions for interventions. As clinical data are complex and varied in their structure and components, continued innovation of modeling strategies is required to identify architectures that can best model outcomes. In this work, we train a Heterogeneous Graph Model (HGM) on Electronic Health Record data and use the resulting embedding vector as additional information added to a Convolutional Neural Network (CNN) model for predicting in-hospital mortality. We show that the additional information provided by including time as a vector in the embedding captures the relationships between medical concepts, lab tests, and diagnoses, which enhances predictive performance. We find that adding HGM to a CNN model increases the mortality prediction accuracy up to 4\%. This framework serves as a foundation for future experiments involving different EHR data types on important healthcare prediction tasks.
\end{abstract}


\end{frontmatter}

\section{Introduction}
Timely prediction of in-hospital mortality within intensive care units (ICU) is beneficial \cite{sharma2017mortality,johnson2017real} for practitioners to tailor care and allow for earlier interventions to prevent deterioration \cite{delahanty2019development, meyer2018machine}. Electronic Health Record (EHR) data consist of information relating to patient encounters with a health system, such as demographics, disease diagnoses, vital signs, and medications, among others \cite{jensen2012mining, glicksberg2018next} which are often used for machine learning (ML) predictions for different tasks in the biomedical domain including mortality prediction  \cite{rajkomar2018scalable,shickel2017deep,glicksberg2018automated}.  The inherent complexity of EHR data often require advanced modeling frameworks to gain robust performance for these tasks. A common modeling approach for EHR research is a 2-dimensional convolutional neural networks (CNN) with one dimension as time and the other as clinical features \cite{zhang2017hcnn,kim2019deep,cheng2016risk}. In healthcare-related CNN models, various medical features are normally concatenated to be directly used as inputs and create embeddings \cite{miotto2016deep,de2020phe2vec,Landi2020}. This form of feature representation can be powerful, but disregards the graphical structure and interconnectivity between medical concepts \cite{choi2019graph,choi2018mime} which can affect the CNN performance especially since EHR data is often sparse due to missingness \cite{cheng2016risk}.

In this work, we propose a Heterogeneous Graph Model (HGM) to create a patient embedding vector, which better accounts for missingness in data for training a CNN model. The HGM model captures the relationships between different medical concept types (e.g., diagnoses and lab tests) due to its graphical structure. This relational representation facilitates capturing more complex patient patterns and encoding similarities. 

\section{Methodology}

\subsection{Dataset}
We conduct our experiments on de-identified EHR data from MIMIC-III \cite{johnson2016mimic}. This data set contains various clinical data relating to patient admission to ICU, such as demographics, lab test results, and disease diagnoses. We collected data for 5,956 patients, extracting lab tests every hour from admission. There are a total of 409 unique lab tests and 3,387 unique disease diagnoses observed. We bin the lab test events into 6, 12, 24, and 48 hours prior to patient death or discharge from ICU. From these data, we perform mortality predictions that are 10-fold, cross validated.

\subsection{Convolutional Neural Network Model}
\label{cnn}
CNNs are often used, and perform well, on image processing tasks \cite{krizhevsky2012imagenet} due to their inherent feature extraction and abstraction ability, which increases the accuracy for classification tasks. There are also studies that have demonstrated encouraging successes in using CNN for EHR analyses.  In this work, we use an standard CNN model as the baseline.

Since CNNs typically require two dimensional inputs, we treat time as the horizontal dimension and medical events as the vertical dimension. For the time dimension, we record every event with one-hour binned increments with respect to the patient death or discharge time. In this model, the vertical dimension is constructed by concatenating two medical event vectors: lab tests and diagnoses. Every entry of the lab test vector records the value of a specific lab test by hour. For the diagnosis vector, the i-th entry is 1 if the i-th diagnosis is observed, otherwise 0. We treat mortality prediction as a binary classification, for which we use a softmax layer with two dimensions and cross-entropy for loss.

 \begin{figure*}[h]
    \centering
    \includegraphics[scale=0.34]{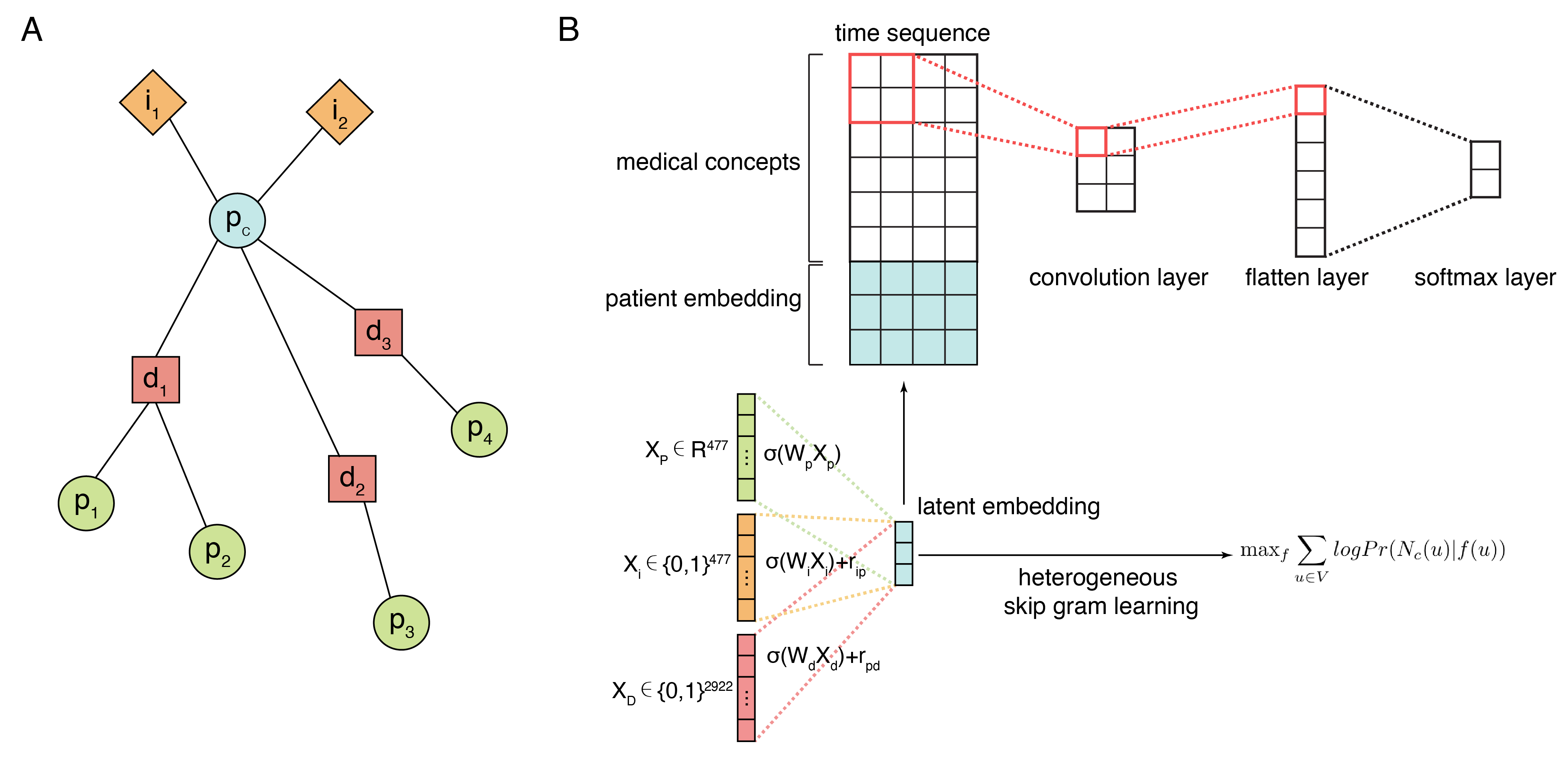}
    \caption{(A) A graphical representation of the HGM for p: patient, i: lab test, and d: diagnosis data. (B)  All graph nodes in (A) have a corresponding vector like those shown in (B).  The vector representations can be projected into a shared space with the TransE method, and this projection is optimized for retaining relations in the original data in the embedding via skip-gram optimization. Finally, these vectors are concatenated into the CNN model for mortality prediction.}
    \label{fig:combine_model}
\end{figure*}

\subsection{Heterogeneous Graph Model}
The features used in baseline CNN model are essentially raw data concatenated together, which does not consider the relationships between medical concepts. We use an HGM to capture these inherent relationships by creating three different type of nodes: patient, lab test, and diagnosis. These different types of nodes are connected by two relation types: tested and diagnosed. These could be represented with two triples:
\begin{align*}
 & Patient \xrightarrow{tested} lab :\{patient, tested, lab\}\\ 
 & Patient \xrightarrow{diagnosed} Diagnosis:\{patient, diagnosed, diagnosis\}\\
\end{align*}
the testing relationship shows whether a specific lab test was given to a patient at a specific time, the diagnosed relationship shows whether a patient was diagnosed with a disease.

To represent the lab test and diagnosis node types, we use multi-hot encoding vector: $X_l \in \{0,1\}^{409}$ and $X_d \in \{0,1\}^{3387}$, the i-th entry with the value of 1 indicates whether a specific lab test was performed or a specific diagnosis was given. 

\subsubsection{Node Embeddings}
For capturing the relations between different medical events related to a patient, we utilize the TransE model\cite{bordes2013translating} to project different type of nodes into the same latent space, then classify those nodes that are connected as a similar group and the disconnected nodes as a dissimilar group. 

The TransE model uses a set of 1) projection matrices and 2) relation vectors. After initialization, projections and translations are optimized end-to-end. Heterogeneous nodes $X_p, X_l, X_d$ are projected into a shared latent space with trainable projection matrices $W_{p},W_{i},W_{d}$ using the nonlinear mappings:
\begin{equation}
\begin{split}
    c_{p}&=\sigma(W_{p}\cdot X_{p})\\
    c_{i}&=\sigma(W_{i}\cdot X_{i})\\
    c_{d}&=\sigma(W_{d}\cdot X_{d})
\end{split}
\end{equation}
Where $\sigma$ is a non-linear activation function and $c_{p},c_{i},c_{d}$ are the latent representations of each type of node. Despite the fact that the EHR uses different dimensions for different data types $X_{p},X_{i},X_{d}$, all nodes types are projected into the same latent space. Then we apply translation operations to link these different types of nodes:

\begin{equation}
\begin{split}
    c_{p}&=c_{i}+r_{ip}\\
    c_{d}&=c_{p}+r_{pd}\\
\end{split}
\end{equation}

where $r_{ip}$ and $r_{pd}$ are the relation vectors connecting patients to lab tests and diagnoses, respectively. Both $c_{p}^{'}$ and $c_{p}$ use the same projection matrix $W_p$.

\subsubsection{Optimization Model}
For training the HGM, we apply a skip-gram optimization model~\cite{dong2017metapath2vec},which increases the proximity between embedding points whose corresponding graph nodes are often connected after the projection and translation operations:
\begin{equation}
    \mathrm{max}\sum_{u\in V}\sum_{t\in T_{V}}logPr(N_{t}(u)|u)
    \label{hetero}
\end{equation}
where $N_{t}(u)$ are the neighborhood vertices of center node $u$, and $t\in T_{V}$ is the node type. Here, we learn the node embeddings by maximizing the probability of correctly predicting the the patient node's associated lab tests and diagnoses. The prediction probability is modeled as a softmax function:
\begin{equation}
        Pr(c_{t}|f(u))=\frac{e^{\vec{c}_{t}\cdot \vec{u}}}{Z_{u}}%\sum_{j}e^{c_{j}^{T}x_{i}}}
    \label{soft_max}
\end{equation}
% $u$ corresponds to X_p from above
where $\vec{u}$ is the latent representation of patient $u$, $\vec{c}_{t}$ is the latent representation of lab and diagnosis neighbors of node $u$, and $\vec{c}_{t}\cdot \vec{u}$ is the inner product of the two embedding vectors representing their similarity. $Z_{u}$ is the normalization term $Z_{u} = \sum_{v\in V}e^{\vec{v}_{t}\cdot \vec{u}}$ that is a sum over all vertices $V$, each of which is represented as $\vec{v}_{t}$ including all node types. Therefore, equation \ref{hetero} is simplified to:
\begin{equation}
    \mathcal{L}_{s}=-\sum_{t\in T}\sum_{u\in V}\Big[\sum_{c_{t}\in N_{t}(u)}\vec{c_{t}}\cdot \vec{u}-logZ_{u}\Big]
    \label{sim_loss}
\end{equation}

Numerical computation of $Z_{u}$ is intractable for large-scale graphs. So we adopt negative sampling strategy~\cite{mikolov2013distributed} to approximate the normalization factor. We eventually use the following optimization function:
%\begin{equation}
\begin{align}
     \mathcal{L}_{s}&=-\sum_{t\in T}\sum_{u\in V}\Big[\sum_{c_{t}\in N_{t}(u)}log\sigma(\vec{c_{t}}\cdot \vec{u})+
     \sum_{j=1}^{\mathbb{K}}E_{c_{j}\sim P_{v}(c_{j})}log\sigma(-\vec{c_{j}}\cdot \vec{u})\Big]
\label{SGNN_loss}
\end{align}
%\end{equation}

where $\sigma(x)=\frac{1}{1+\exp(-x)}$, $\mathbb{K}$ is the number of negative samples. $P_{v}(c_{j})$ is the negative sampling distribution.\\

% \subsubsection{Training Specifics}
 For training HGM, we perform heterogeneous neighborhood sampling by its one-hop connectivity, and pick $Patient$ node as the center node, since it has one-hop connections to both $Diagnoses$ and $Lab\_test$ nodes. Specifically, for one training center $Patient$ node, we uniformly sampled 10 $Diagnoses$ one-hop direct connected nodes, and 10 $Lab\_test$ one-hop direct connected nodes. From these sampled 10 $Diagnoses$ nodes, we sample another 10 $Patient$ nodes, each having connections with each of the prior 10 $Diagnoses$ nodes. In this way, we connect the center patient node with similar other $Patient$ nodes by their common diagnoses. We also sample the patient node which belongs to the next hour corresponding to the center $Patient$ node. For negative sampling~\cite{mikolov2013distributed}, we perform uniform sampling through all $Diagnoses$ node and $Lab\_test$ nodes that do not have one-hop connections with the center training patient node. We then project these different nodes into same latent space through TransE model.  After unifying the embeddings for different node types, each concept is represented as a point in a Euclidean space.  In this space, we can measure the similarity between any two vectors using dot product.

 \subsection{HGM Embeddings with CNN Model}
 The HGM embedding vector encodes not only a patient's information, but also their relation with diagnoses, lab tests, and subsequent lab test results in time. The patient node is represented as a vector $X_p \in \mathbb{R}^{477}$ containing the numerical values measured from lab tests averaged at that time step. We concatenate the resulting embedding vectors to feed into the baseline CNN vertical feature dimension to form a final feature vector within every hour, and use these new features as the CNN input to predict mortality. In addition, since we encode time as a relation type, we can infer the embedding vector of time steps with missing data based on information from the previous hour. We visualize this procedure in Fig 1.

\section{Experiments}
\begin{figure}[t]
    \centering
    \includegraphics[width=\textwidth]{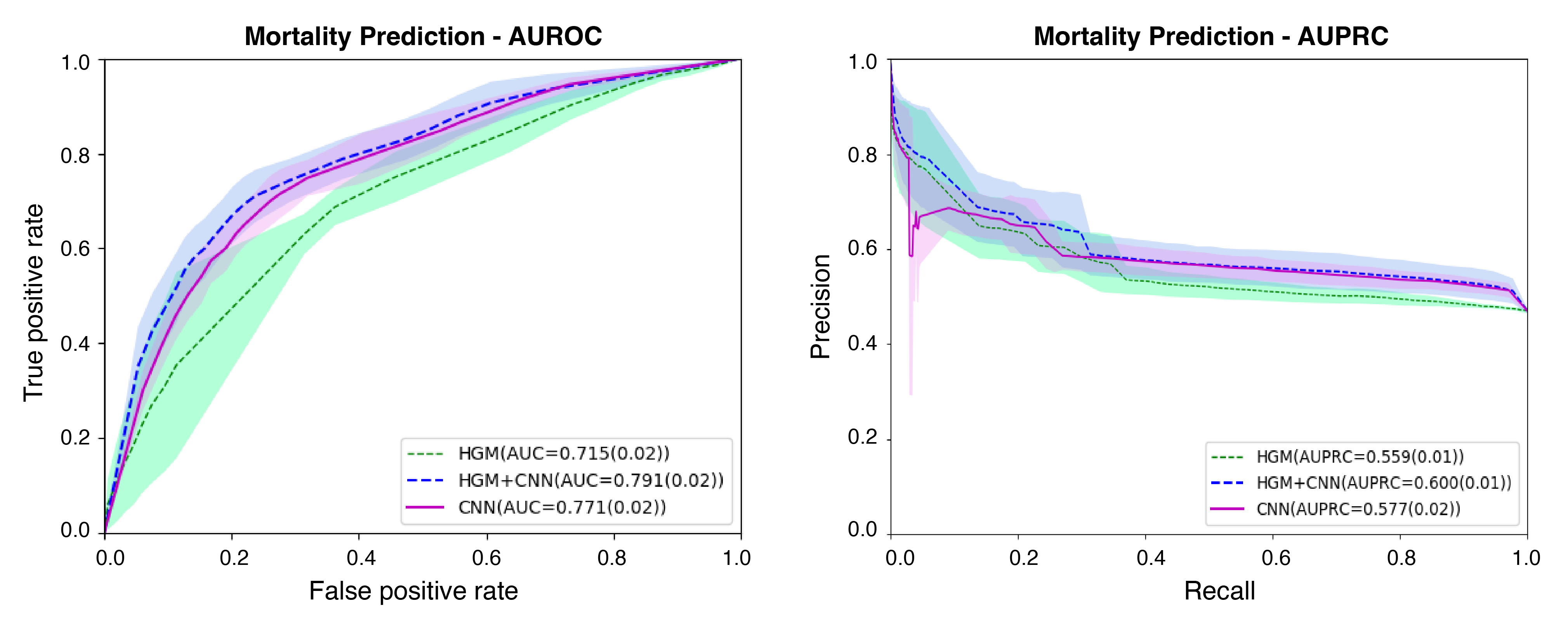}
    \caption{Evaluation of AUROC and AUPRC curves for HGM, CNN, and HGM+CNN models}
    \label{fig:roc_curve}
\end{figure}

We aim to predict mortality 6, 12, 24, and 48 hours prior to death and/or discharge. The CNN model is used for prediction as introduced in section \ref{cnn}. We compare three different scenarios to test the impact of adding HGM embedding vectors as additional features to the framework:\\
\begin{itemize}
\item HGM: Embed patient labs and  diagnosis raw data.
\item CNN: Use raw lab test feature.
\item {HGM+CNN}: Concatenate the HGM patient embedding vector, and the raw lab test feature vector.
\end{itemize}
In this work, we use AUROC and AUPRC scores as the primary performance metric. We tabulate the results in Table\ref{tab:accur} and we show the evaluation AUROC and AUPRC curves for these tasks in Fig.\ref{fig:roc_curve}

\begin{table}[t]
  \centering
  \caption{Mortality prediction AUROC evaluation. Mean values from 10-fold cross validation with standard deviation for confidence intervals.}
    \begin{tabular}{p{2cm}p{2cm}p{2cm}p{1.5cm}}
    \toprule
   % \multicolumn{1}{l}{\multirow{}{}{Hours Prior to Death}} &
    %\multicolumn{1}{l}{\multirow{}{}{Hours Prior to Death}} &
    %\multicolumn{3}{c}{Models} \\
%          & %\multicolumn{1}{l}{HGM} & \multicolumn{1}{l}{CNN} & \multicolumn{1}{l}{HGM+CNN} \\
    \multicolumn{1}{l}{\multirow{2}[0]{*}{Hours prior to death}} & \multicolumn{3}{c}{Models} \\
    \cmidrule{2-4}
    & HGM     & CNN     & HGM+CNN \\
    \midrule
     6     & 0.714$\pm{0.02}$      & 0.782$\pm{0.01}$      &\textbf{0.800}$\pm{0.01}$  \\
    12    & 0.715$\pm{0.03}$      & 0.771$\pm{0.02}$     & \textbf{0.791}$\pm{0.02}$ \\
    24    & 0.653$\pm{0.03}$      &0.775$\pm{0.01}$       & \textbf{0.796}$\pm{0.01}$ \\
    48    &0.641$\pm{0.03}$      &0.767$\pm{0.01}$     & \textbf{0.771}$\pm{0.01}$\\
    \bottomrule
    \end{tabular}%
  \label{tab:accur}%
\end{table}

\begin{table}[t]
  \centering
  \caption{Mortality prediction Mean values from 10-cross validation with standard deviation for confidence intervals.}
    \begin{tabular}{p{2cm}p{2cm}p{2cm}p{1.5cm}}
    \toprule
   % \multicolumn{1}{l}{\multirow{}{}{Hours Prior to Death}} &
    %\multicolumn{1}{l}{\multirow{}{}{Hours Prior to Death}} &
    %\multicolumn{3}{c}{Models} \\
%          & %\multicolumn{1}{l}{HGM} & \multicolumn{1}{l}{CNN} & \multicolumn{1}{l}{HGM+CNN} \\
    \multicolumn{1}{l}{\multirow{2}[0]{*}{Hours prior to death}} & \multicolumn{3}{c}{Models} \\
    \cmidrule{2-4}
    & HGM     & CNN     & HGM+CNN \\
    \midrule
     6     & 0.557$\pm{0.02}$      & 0.590$\pm{0.01}$      &\textbf{0.601}$\pm{0.01}$  \\
    12    & 0.559$\pm{0.02}$      & 0.577$\pm{0.02}$     & \textbf{0.600}$\pm{0.01}$ \\
    24    & 0.578$\pm{0.02}$      &0.589$\pm{0.01}$       & \textbf{0.604}$\pm{0.01}$ \\
    48    &0.567$\pm{0.03}$      &0.585$\pm{0.02}$     & \textbf{0.617}$\pm{0.02}$\\
    \bottomrule
    \end{tabular}%
  \label{tab:accur}%
\end{table}

The testing results shows that the HGM+CNN outperforms both the basic HGM and CNN models, indicating the additional information added from the HGM patient embeddings increase the accuracy of predicting in-patient mortality. The prediction accuracy of using different hours prior to death and/or discharge does not vary by much, indicating that different time windows do not have a major impact on the result for this particular task and modeling strategy. The prediction accuracy in the CNN model drops by 1\% in the case of six hours prior to death and/or discharge, but not in the other two models, indicating that using the embedding features from HGM model is slightly more robust than the raw data.

\section{Discussion and Conclusion}
In this work, we propose a method to incorporate patient embedding vector from a HGM model into a CNN model in order to provide more information via interconnectivity between different clinical concepts. We assess the value of this implementation on a task of predicting mortality in EHR data. The results of our experiment shows the superior performance of adding the additional patient embedding vector, which is pretrained from the HGM model, compared to pure raw features as the input to CNN model. In one aspect, this is due to the fact that the HGM embedding vector captures additional relational information between different medical concepts, thus providing additional information to CNN model. 

Furthermore, we observe that concatenating the HGM embedding vector with diagnosis feature vectors does not increase the accuracy versus using the concatenation between raw lab test and diagnosis feature vectors. This finding indicates that the raw lab test feature vector can provide unique information for CNN to utilize. At the same time, this finding indicates that the embedded patient vector from HGM model could lose some information from the raw lab test feature along the process of projecting these data into a low dimensional latent space. By concatenating all feature vectors, we aim to preserve the information from different data points, which helps to achieve higher mortality prediction accuracy. We hope the findings from this work can be expanded in future directions that may add more EHR node types and time components on a variety of other important health-related predictive tasks.



\section*{Author Contributions}

TW designed the study and performed the analyses. TW and BSG wrote the manuscript. TW, HH, AA, YD, and BSG evaluated the results and edited the manuscript. YD, AA, and BSG supervised the project.

\section*{Acknowledgements}

\bibliographystyle{ACM-Reference-Format}
\bibliography{ref}

\end{document}